\documentclass[sigconf, nonacm]{acmart}

\usepackage{makecell}

\AtBeginDocument{%
  }

\begin{document}

\title{MMPKUBase: A Comprehensive and High-quality Chinese Multi-modal Knowledge Graph}

\author{Xuan Yi}
\affiliation{%
  \institution{Wangxuan Institute of Computer Technology, Peking University}
  \city{Beijing}
  \country{China}}
\email{yixuan2020@stu.pku.edu.cn}

\author{Yanzeng Li}
\affiliation{%
  \institution{Wangxuan Institute of Computer Technology, Peking University}
  \city{Beijing}
  \country{China}}
\email{liyanzeng@stu.pku.edu.cn}

\author{Lei Zou}
\affiliation{%
  \institution{Wangxuan Institute of Computer Technology, Peking University}
  \city{Beijing}
  \country{China}}
\email{zoulei@pku.edu.cn}

\begin{abstract}
Multi-modal knowledge graphs have emerged as a powerful approach for information representation, combining data from different modalities such as text, images, and videos. While several such graphs have been constructed and have played important roles in applications like visual question answering and recommendation systems, challenges persist in their development. These include the scarcity of high-quality Chinese knowledge graphs and limited domain coverage in existing multi-modal knowledge graphs. This paper introduces MMPKUBase, a robust and extensive Chinese multi-modal knowledge graph that covers diverse domains, including birds, mammals, ferns, and more, comprising over 50,000 entities and over 1 million filtered images. To ensure data quality, we employ Prototypical Contrastive Learning and the Isolation Forest algorithm to refine the image data. Additionally, we have developed a user-friendly platform to facilitate image attribute exploration.
\end{abstract}


\maketitle

\section{Introduction}
Knowledge graphs are a crucial method for organizing and representing connected information. A traditional knowledge graph is usually a text-based network, including entities with textual content and relationships between them. Major examples include Wikidata \cite{vrandevcic2014wikidata}, DBpedia \cite{auer2007dbpedia}, Freebase \cite{bollacker2008freebase}, and YAGO \cite{suchanek2007yago}. These knowledge graphs are excel at describing semantic relationships and supporting various downstream applications. However, they do not integrate and utilize multi-modal information. As multi-modal data continues to grow, this limitation is becoming increasingly apparent.

In recent years, multi-modal knowledge graphs has gained significant attention. These graphs integrate various modalities like text, images, and videos to form comprehensive knowledge representations. They play a vital role in various tasks, including visual question answering systems \cite{yu2020cross}, recommendation engines \cite{sun2020multi,tao2021multi}, etc. While the research community has made substantial progress in constructing high-quality multi-modal knowledge graphs \cite{li2020gaia,wen2021resin,ma2022mmekg,onoro2017imagegraph,liu2019mmkg,wang2020richpedia,alberts2020visualsem}, several challenges and limitations persist: 

Despite the abundance of knowledge graphs in English and other languages, there is a lack of high-quality Chinese knowledge graphs. This shortage limits the use of multi-modal knowledge graphs in the Chinese-speaking world. Furthermore, many existing multi-modal knowledge graphs fall short in terms of domain coverage. Quality is yet another critical issue: the images incorporated into multi-modal knowledge graphs often exhibit disparities in terms of quality and accuracy. 

In this paper, we present MMPKUBase, a comprehensive Chinese multi-modal knowledge graph, with an effective and robust construction methodology.

The main contributions of this paper are as follows:
\begin{itemize}

\item We introduce MMPKUBase, a extensive multi-modal knowledge graph presented in the Chinese language, featuring over 52,180 entities and 1,542,894 images across varied domains including birds, mammals, ferns, monocotyledons, Rosales plants, architecture, archaeological sites, automobiles and military.

\item We use Prototypical Contrastive Learning \cite{li2020prototypical} to extract image features and refine the data by implementing the Isolation Forest algorithm \cite{liu2008isolation}, retaining 1,227,013 high-quality images. These images are not only representative but also suitable for downstream tasks.

\item We develop an user-friendly demonstration platform for easy interaction with the knowledge graph and exploration of image attributes.
\end{itemize}

\section{Related Work} \label{Related Work}

With the development of artificial intelligence technology and the rapid growth of multi-modal data, the construction of high-quality multi-modal knowledge graphs has made significant progress.

The process of constructing multi-modal knowledge graphs involves two main approaches: developing new multi-modal knowledge graphs based on multi-modal documents~\cite{li2020gaia,wen2021resin,ma2022mmekg}, and collecting and selecting relevant images for entities within existing traditional knowledge graphs~\cite{onoro2017imagegraph,liu2019mmkg,wang2020richpedia,alberts2020visualsem}.

In the first approach, a sophisticated system for extracting knowledge is typically established, consisting of branches for text and visual information extraction, followed by the fusion of textual and visual data. GAIA~\cite{li2020gaia} encompasses branches for extracting text-based knowledge, visual knowledge, and modules for fusing cross-media knowledge. Both branches make use of the same types in the DARPA AIDA ontology, using the same multi-modal document set as input for integration within the cross-media knowledge fusion module. Resin \cite{wen2021resin}, much like GAIA, utilizes both text and visual branches, merging them through the coreference of visual and textual elements. Furthermore, a module for schema matching is leveraged to align the information extracted with suitable schema from a schema repository. MMEKG~\cite{ma2022mmekg}, drawing inspiration from GAIA and RESIN knowledge extracting systems, extends and enhances the framework. It replaces all Cross-encoder with Bi-encoder and employs multi-task joint model training for event relation extraction to improve efficiency.

In the second approach, the primary focus is on searching for pertinent images of entities within existing knowledge graphs from sources like search engines, followed by quality control. ImageGraph \cite{onoro2017imagegraph} uses entities from the knowledge graph as query strings to retrieve images from search engines, treating these images as attributes values of the entities. MMKG \cite{liu2019mmkg} extends this method to multiple knowledge graphs, applying rules to eliminate damaged, low-quality, and duplicated images to maintain the quality of the data. To obtain both highly pertinent and varied images for corresponding textual entities, Richpedia \cite{wang2020richpedia} extracts visual feature vectors using VGG16 and reduces dimensionality, applying k-means and cosine similarity for selection. Additionally, Richpedia utilizes related hyperlinks and text from Wikipedia to discover relationships between image entities. VisualSem \cite{alberts2020visualsem}, acknowledging that many entities do not require visual understanding, initiates the process with high-quality visual entities from the multilingual and multi-modal resource BabelNet and explores other related visual entities.

In the context of this paper, the focus aligns with the second approach. The study primarily involves the augmentation of the existing Chinese knowledge graph PKUBase with images obtained from search engines, spanning a diverse array of domains. A novel method for image quality control is introduced to ensure the reliability and relevance of the incorporated images.
\section{Method} \label{Method}

This section first establishes the definition of MMPKUBase. This is followed by an overview of the creation pipeline for MMPKUBase. Then, the intricate details of each component will be explored in depth.

\subsection{Definition}

The key distinction between a traditional knowledge graph and MMPKUBase in this work lies in the nature of attribute values, with traditional KGs using single-modal attribute values, while the multi-modal knowledge graph incorporate multi-modal attribute values to represent information in various forms, such as text and images.

In this work, a multi-modal knowledge graph is denoted as $ \mathcal{G}_{mm} = \{\mathcal{E}, \mathcal{R}, \mathcal{A}, \mathcal{V}_{mm}, \mathcal{T}\}$, where $ \mathcal{E}, \mathcal{R}, \mathcal{A}, \mathcal{V}_{mm}$ represent the set of entities, relationships, attributes and attribute values, respectively. A triple in the knowledge graph can be an element in $ \mathcal{E} \times \mathcal{R} \times \mathcal{E}$ or $ \mathcal{E} \times \mathcal{A} \times \mathcal{V}_{mm}$. \(\mathcal{T}\) represents the set of these triples. The set of attribute values \(\mathcal{V}_{mm}\) is multi-modal, including a subset of text attributes \(\mathcal{V}_{text}\) and a subset of image attributes \(\mathcal{V}_{image}\). For example, the knowledge graph might include triples like <Eiffel Tower, Height, 300m>, where the attribute value is textual, or <Eiffel Tower, Has Image, [Image of Eiffel Tower]>, where the attribute value is an image.

\subsection{Framework Overview}

\begin{figure}[htb]
\includegraphics[width=\columnwidth]{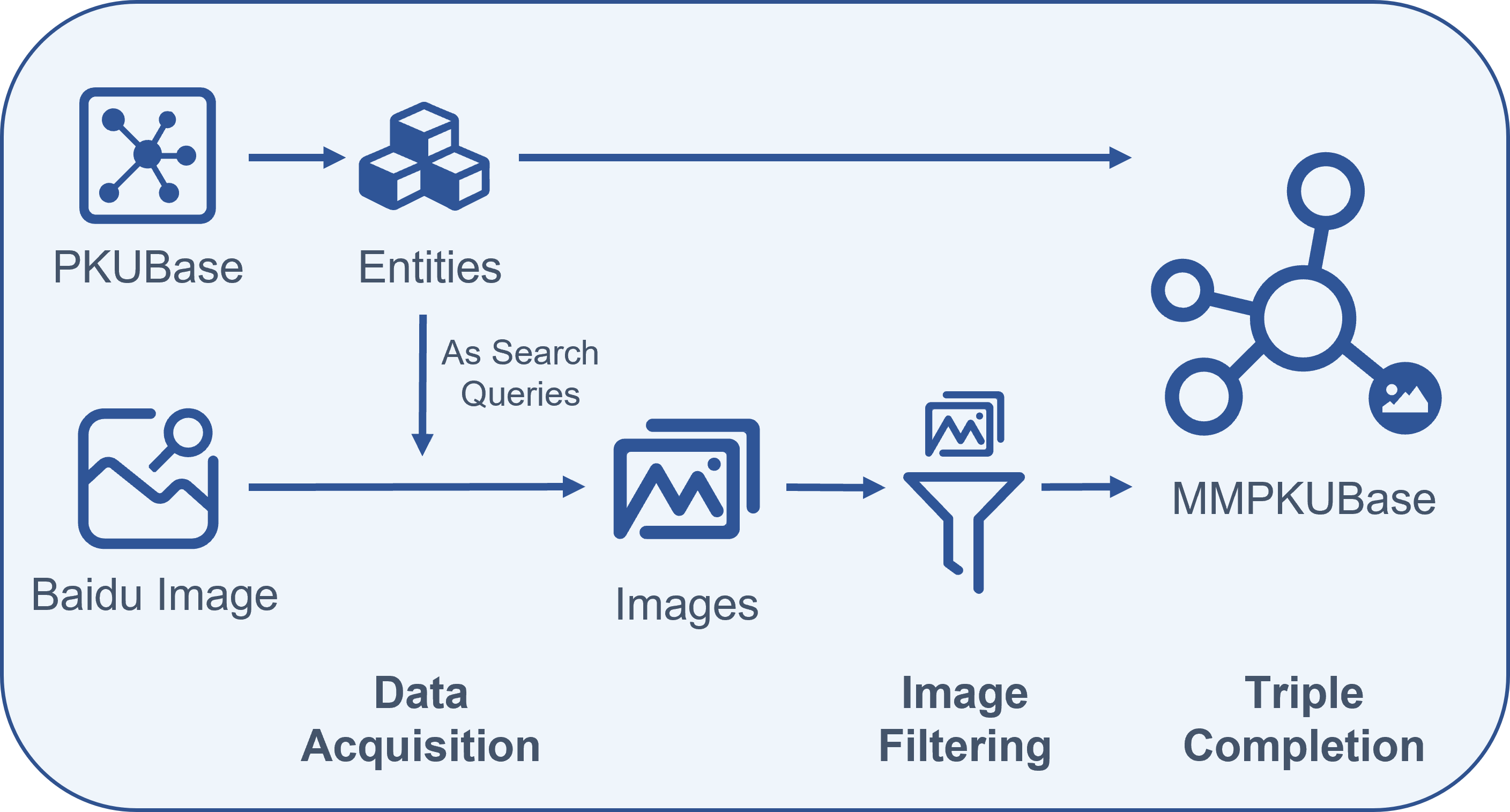} 
\caption{The overview construction pipeline of MMPKUBase}
\label{fig.1}
\end{figure}

Figure \ref{fig.1} illustrates the overview construction pipeline of MMPKUBase, including data acquisition, image filtering and triple completion. The data acquisition phase involves the selection of entities from PKUBase and the collection of pertinent visual resources from search engine. Feature extraction and image quality control processes ensure the purity of the data repository. In the triple completion phase, images are treated as attribute values and stored in the knowledge graph.

\subsection{Data Acquisition}

The process of data acquisition is subdivided into two phases: entity selection and image retrieval. First, the sources of data will be clarified:

\subsubsection{Data Sources}
The data to build this knowledge graph comes from two data sources: PKUBase\footnote{\url{http://pkubase.gstore.cn}} and Baidu Image\footnote{\url{https://image.baidu.com/}}.

\begin{itemize}

\item \textbf{PKUBase} is a massive Chinese structured knowledge resource constructed from semi-structured and unstructured trusted texts through a series of natural language processing tools and machine learning methods, containing a standard Chinese category system framework, nearly 10 million Chinese entities, and more than 60 million Chinese knowledge entries. Its comprehensive and extensive data making it an ideal source of textual information for our project.
\item \textbf{Baidu Image} provides a reliable image search service based on all kinds of pictures extracted from billions of Chinese web pages, and is widely used in Chinese communities. So far, the Baidu image search engine can retrieve nearly one hundred million images. It will be used to acquire images associated with the entities in our knowledge graph.

\end{itemize}
\subsubsection{Entity Selection}

In this phase, entities with clear visual attributes are prioritized in our project to enhance performance in visually-oriented queries and applications, while entities without direct visual representations are excluded to optimize resource use. Consequently, nine domains have been selected based on their distinct visual characteristics: birds, mammals, ferns, monocotyledons, Rosales, architecture, archaeological sites, automobiles and military.

After selecting the domain, the entity selection process was executed through a combination of automated and manual methods.

In PKUBase, entities are associated with a predicate ``type'', which we utilized to filter relevant entities for different domains. For domains of architecture, archaeological sites, automobiles, and military, entities are filtered using the triplet <entity, type, domain name>. For the animal and plant domains, the filtering process began with <entity name, type, animal/plant>, followed by a refinement based on biological taxonomy. For instance, entities in the Birds domain are selected from those that are categorized as animals and meet the <Entity, Category, Aves> criteria. After the automatic filtering, entities unrelated to the intended domains, such as humans or books, were excluded using a manually curated lexicon.

\subsubsection{Image Retrieval}
Once the entities are selected, the next step is to gather relevant visual data. This is achieved through a process of image crawling, whereby the chosen entities' names serve as queries for image searches on the Baidu Image. For each selected entity, up to 30 of the most relevant images are crawled from the search results. If the search results yield fewer than 30 images, all available images are crawled. The number was chosen to ensure that the multi-modal knowledge graph is enriched with a diverse and comprehensive set of visual data directly related to the selected entities.

In summation, our data acquisition process has yielded a total of 52,180 selected entities and 1,539,894 related images.

\subsection{Image Filtering}
This stage, the image filtering process, demands rigorous screening to exclude images that could potentially compromise the quality of MMPKUBase. Firstly, corrupted images or those in animated formats are removed, resulting in a remaining set of 1,535,005 images. Secondly, images that exhibit little relevance to the entities are filtered out as outliers. To achieve this, we utilize Prototypical Contrastive Learning (PCL) \cite{li2020prototypical} for image feature extraction and apply Isolation Forest \cite{liu2008isolation} for filtering.
\subsubsection{PCL Feature Generation}
This phase focus on the acquisition of image features, employing Prototypical Contrastive Learning as a key technique.

Prototypical Contrastive Learning (PCL) represents a unsupervised representation learning approach that amalgamates the principles of contrastive learning and clustering. In PCL, prototypes are defined as representative embeddings of a group of semantically similar instances. An instance can belong to multiple prototypes of varying granularity. PCL learn an embedding space that make samples more similar to the prototype they are associated with, compared to other prototypes. In the context of our work, the images collected from various thematic domains exhibit features that align with this concept. For instance, images of the entity `mandarin duck' and `cotton pygmy goose' can be associated with medium-grained prototypes unique to their respective categories, while simultaneously belonging to a higher-level `Anatidae' prototype. Furthermore, within the image collection of the entity `mandarin duck', distinct prototypes can represent the male and female birds, reflecting finer granularity.

For each thematic domain, we individually train PCL models (with ResNet50\cite{He2016resnet} as the backbone, running for 100 epochs), which subsequently yield 128-dimensional image representations. These representations preserve the distinctive visual traits and semantic nuances of each image.

\subsubsection{Image Selection}
\begin{figure}[htb]
\begin{center}
\includegraphics[width=\columnwidth]{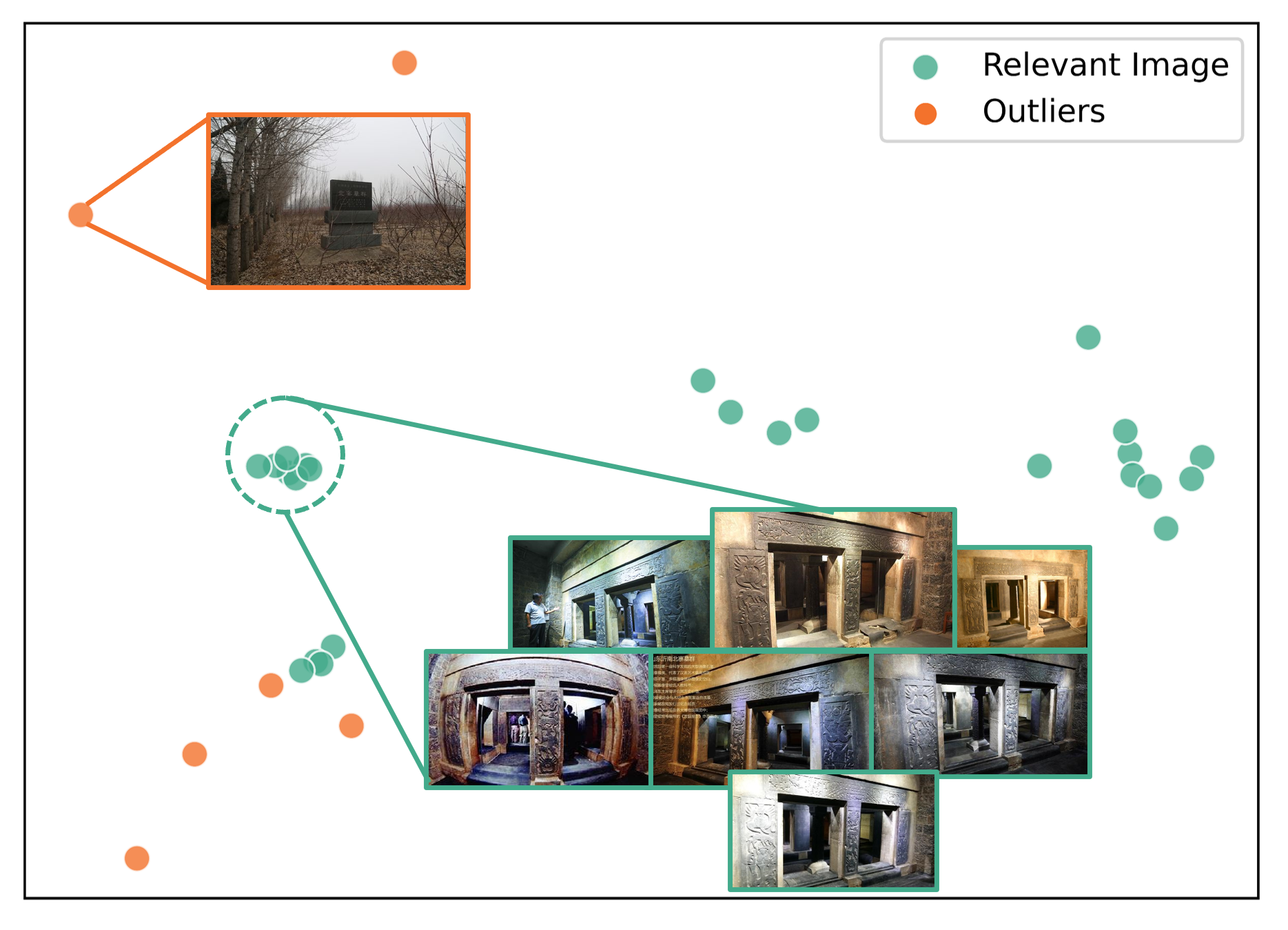} 
\caption{The results of filtering a collection of images for a specific architectural entity using the Isolation Forest method.}
\label{fig.2}
\end{center}
\end{figure}

Our method entails a selection process predicated on the similarity of images within the search results. It is assumed that the majority of the search results are sufficiently relevant to the entity. Consequently, it is expected that, for each entity's search results, relevant images will aggregate into several clusters, while irrelevant search outcomes will be identified as anomalies, distinct and distant from the clusters.

To effectively identify and isolate these irrelevant or noisy images, the Isolation Forest technique, a robust outlier detection method in high-dimensional datasets, is employed. This approach is applied to the PCL features extracted in the previous step for each entity's search results. The `contamination' parameter, representing the proportion of outliers within the dataset, is configured at 0.2. To exemplify our approach and visualize the clusters and outliers, Figure \ref{fig.2} provides an example. In this filtering process, images of frequently occurring architectural structures cluster together, reflecting their high relevance, while images with lower relevance and infrequent occurrences are identified as outliers. UMAP (Uniform Manifold Approximation and Projection) \cite{mcinnes2018umap} is employed to reduce dimension for visualization purposes.

\subsection{Triple Completion}
In the final phase of our methodology, the curated and filtered images are connected to entities as multi-modal attributes in the set $\mathcal{V}_{mm}$. Triples in $ \mathcal{E} \times \mathcal{A} \times \mathcal{V}_{mm}$ , of the form <entity, with image, [image of entity]>, are subsequently incorporated into the multi-modal knowledge graph, following the RDF format\footnote{\url{https://www.w3.org/RDF/}}.

With the completion of this phase, the construction of MMPKUBase, our multi-modal knowledge graph, is finalized.

\begin{figure*}[htb]
\begin{center}
\includegraphics[scale=0.73]{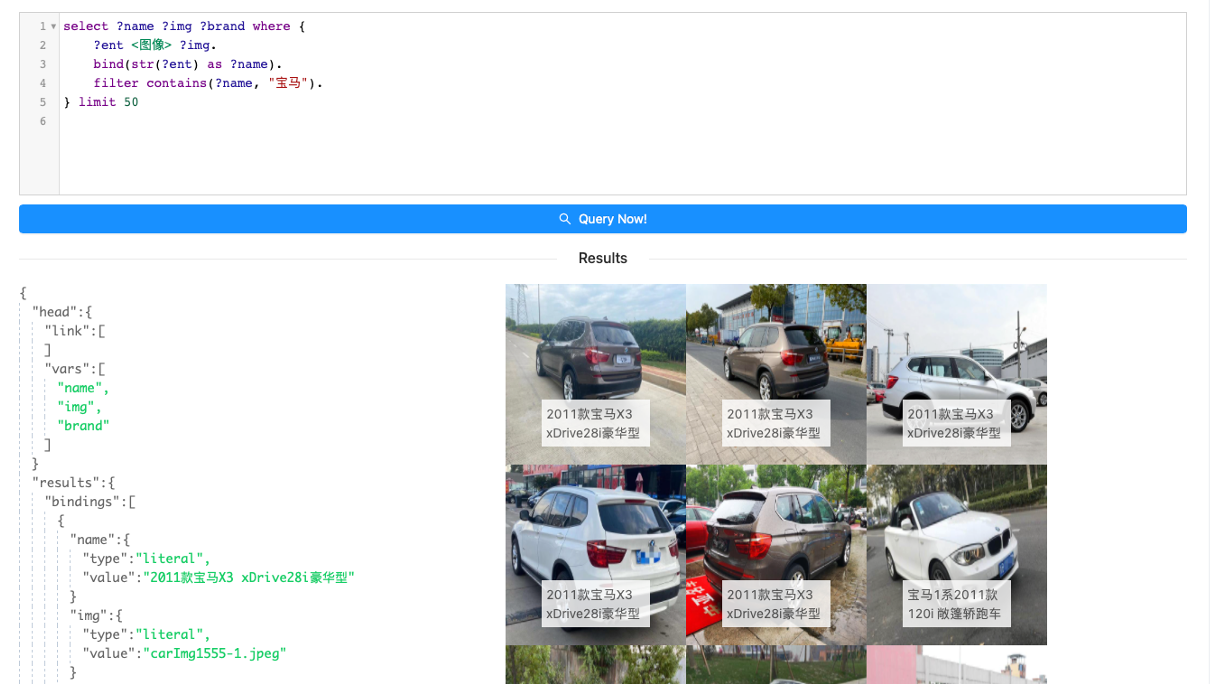} 
\caption{Query examples from the demonstration platform. The SPARQL query is designed to locate entities whose names contain the substring `BMW' and retrieve their associated image attributes.}
\label{fig.3}
\end{center}
\end{figure*}

\section{Statistics} \label{Statistics}

Table \ref{table.1} provides a comprehensive overview of the statistical data pertinent to the development of MMPKUBase, encompassing the number of entities within each thematic domain and the quantity of images both before and after the filtering process. This data plays a pivotal role in assessing the richness of visual content and the level of information available for each thematic domain.

\begin{table}[htb]
\centering
\begin{tabular}{lrrrr}
\hline
\textbf{Topic}&\textbf{Entities}&\textbf{Images}&\textbf{Filtered Images} \\ 
\hline
Birds & 10,554 &  314,977  &  251,326 \\
Mammals & 4,031 &  118,550 &  94,341 \\
Ferns & 2,995 &  90,092 &  69,506 \\
Monocotyledons & 7,822 &  231,570 &  184,479 \\
Rosales Plants & 4,545 &  135,183 &  107,934 \\
Architecture & 3,622 &  106,360 &  84,835 \\
Archaeological Sites & 5,064 &  148,956 &  118,706 \\
Automobiles & 1,911 &  57,139 &  45,666 \\
Military & 11,636 &  340,067 &  270,220 \\
Total & 52,180 &  1,542,894 & 1,227,013 \\
\hline
\end{tabular}
\caption{Entity and Image Count Information.}
\label{table.1}
\end{table}

\section{Demonstration}  \label{Demonstration}

Leveraging the RDF-formatted dataset constructed in the previous sections, we have develop a platform for retrieving and visualizing multi-modal data. This platform enables users to perform SPARQL queries within MMPKUBase, facilitating the retrieval of pertinent entities along with their corresponding visual attributes.

Figures \ref{fig.3} represents a query example. Users can input SPARQL query statements in the search box located at the top of the interface to obtain and explore the multi-modal knowledge graph. The result part is divided into two sections. On the left side, structured query responses are displayed and on the right side, images are presented alongside their corresponding entity names, allowing users to  brows the multi-modal attributes associated with the entities.

\section{Conclusion}  \label{Conclusion}
This paper introduce MMPKUBase, a comprehensive and extensive multi-modal knowledge graph in the Chinese language. MMPKUBase encompasses a diverse range of domains, including birds, mammals, ferns, monocotyledons, Rosales plants, architecture, archaeological sites, automobiles, and military. With over fifty thousand entities and images in the million range, it serves as a valuable resource for various applications. Prototypical Contrastive Learning and the Isolation Forest algorithm is employed to ensure the quality and reliability of our data, resulting in a collection of high-quality and representative images that can be applied to a wide array of downstream tasks. Furthermore, to make MMPKUBase accessible and user-friendly, we developed an intuitive demonstration platform that enables users to query and explore image attributes.

As future work, our focus will be to seamlessly integrate MMPKUBase into real-world applications, unlocking its full potential. Additionally, we are committed to continually enhancing the size and diversity of the knowledge graph, ensuring it encompasses an even broader array of entities and domains.

\bibliographystyle{ACM-Reference-Format}
\bibliography{sample-base}


\begin{thebibliography}{18}

%%% the Web address.

\ifx \showCODEN    \undefined \def \showCODEN     #1{\unskip}     \fi
\ifx \showDOI      \undefined \def \showDOI       #1{#1}\fi
\ifx \showISBNx    \undefined \def \showISBNx     #1{\unskip}     \fi
\ifx \showISBNxiii \undefined \def \showISBNxiii  #1{\unskip}     \fi
\ifx \showISSN     \undefined \def \showISSN      #1{\unskip}     \fi
\ifx \showLCCN     \undefined \def \showLCCN      #1{\unskip}     \fi
\ifx \shownote     \undefined \def \shownote      #1{#1}          \fi
\ifx \showarticletitle \undefined \def \showarticletitle #1{#1}   \fi
\ifx \showURL      \undefined \def \showURL       {\relax}        \fi
% invisible to TeX
\providecommand\bibfield[2]{#2}
\providecommand\bibinfo[2]{#2}
\providecommand\natexlab[1]{#1}
\providecommand\showeprint[2][]{arXiv:#2}

\bibitem[Alberts et~al\mbox{.}(2020)]%
        {alberts2020visualsem}
\bibfield{author}{\bibinfo{person}{Houda Alberts}, \bibinfo{person}{Teresa Huang}, \bibinfo{person}{Yash Deshpande}, \bibinfo{person}{Yibo Liu}, \bibinfo{person}{Kyunghyun Cho}, \bibinfo{person}{Clara Vania}, {and} \bibinfo{person}{Iacer Calixto}.} \bibinfo{year}{2020}\natexlab{}.
\newblock \showarticletitle{Visualsem: a high-quality knowledge graph for vision and language}.
\newblock \bibinfo{journal}{\emph{arXiv preprint arXiv:2008.09150}} (\bibinfo{year}{2020}).
\newblock


\bibitem[Auer et~al\mbox{.}(2007)]%
        {auer2007dbpedia}
\bibfield{author}{\bibinfo{person}{S{\"o}ren Auer}, \bibinfo{person}{Christian Bizer}, \bibinfo{person}{Georgi Kobilarov}, \bibinfo{person}{Jens Lehmann}, \bibinfo{person}{Richard Cyganiak}, {and} \bibinfo{person}{Zachary Ives}.} \bibinfo{year}{2007}\natexlab{}.
\newblock \showarticletitle{Dbpedia: A nucleus for a web of open data}. In \bibinfo{booktitle}{\emph{international semantic web conference}}. Springer, \bibinfo{pages}{722--735}.
\newblock


\bibitem[Bollacker et~al\mbox{.}(2008)]%
        {bollacker2008freebase}
\bibfield{author}{\bibinfo{person}{Kurt Bollacker}, \bibinfo{person}{Colin Evans}, \bibinfo{person}{Praveen Paritosh}, \bibinfo{person}{Tim Sturge}, {and} \bibinfo{person}{Jamie Taylor}.} \bibinfo{year}{2008}\natexlab{}.
\newblock \showarticletitle{Freebase: a collaboratively created graph database for structuring human knowledge}. In \bibinfo{booktitle}{\emph{Proceedings of the 2008 ACM SIGMOD international conference on Management of data}}. \bibinfo{pages}{1247--1250}.
\newblock


\bibitem[He et~al\mbox{.}(2016)]%
        {He2016resnet}
\bibfield{author}{\bibinfo{person}{Kaiming He}, \bibinfo{person}{Xiangyu Zhang}, \bibinfo{person}{Shaoqing Ren}, {and} \bibinfo{person}{Jian Sun}.} \bibinfo{year}{2016}\natexlab{}.
\newblock \showarticletitle{Deep residual learning for image recognition}. In \bibinfo{booktitle}{\emph{Proceedings of the IEEE conference on computer vision and pattern recognition}}. \bibinfo{pages}{770--778}.
\newblock


\bibitem[Li et~al\mbox{.}(2020b)]%
        {li2020prototypical}
\bibfield{author}{\bibinfo{person}{Junnan Li}, \bibinfo{person}{Pan Zhou}, \bibinfo{person}{Caiming Xiong}, {and} \bibinfo{person}{Steven~CH Hoi}.} \bibinfo{year}{2020}\natexlab{b}.
\newblock \showarticletitle{Prototypical contrastive learning of unsupervised representations}.
\newblock \bibinfo{journal}{\emph{arXiv preprint arXiv:2005.04966}} (\bibinfo{year}{2020}).
\newblock


\bibitem[Li et~al\mbox{.}(2020a)]%
        {li2020gaia}
\bibfield{author}{\bibinfo{person}{Manling Li}, \bibinfo{person}{Alireza Zareian}, \bibinfo{person}{Ying Lin}, \bibinfo{person}{Xiaoman Pan}, \bibinfo{person}{Spencer Whitehead}, \bibinfo{person}{Brian Chen}, \bibinfo{person}{Bo Wu}, \bibinfo{person}{Heng Ji}, \bibinfo{person}{Shih-Fu Chang}, \bibinfo{person}{Clare Voss}, {et~al\mbox{.}}} \bibinfo{year}{2020}\natexlab{a}.
\newblock \showarticletitle{Gaia: A fine-grained multimedia knowledge extraction system}. In \bibinfo{booktitle}{\emph{Proceedings of the 58th Annual Meeting of the Association for Computational Linguistics: System Demonstrations}}. \bibinfo{pages}{77--86}.
\newblock


\bibitem[Liu et~al\mbox{.}(2008)]%
        {liu2008isolation}
\bibfield{author}{\bibinfo{person}{Fei~Tony Liu}, \bibinfo{person}{Kai~Ming Ting}, {and} \bibinfo{person}{Zhi-Hua Zhou}.} \bibinfo{year}{2008}\natexlab{}.
\newblock \showarticletitle{Isolation forest}. In \bibinfo{booktitle}{\emph{2008 eighth ieee international conference on data mining}}. IEEE, \bibinfo{pages}{413--422}.
\newblock


\bibitem[Liu et~al\mbox{.}(2019)]%
        {liu2019mmkg}
\bibfield{author}{\bibinfo{person}{Ye Liu}, \bibinfo{person}{Hui Li}, \bibinfo{person}{Alberto Garcia-Duran}, \bibinfo{person}{Mathias Niepert}, \bibinfo{person}{Daniel Onoro-Rubio}, {and} \bibinfo{person}{David~S Rosenblum}.} \bibinfo{year}{2019}\natexlab{}.
\newblock \showarticletitle{MMKG: multi-modal knowledge graphs}. In \bibinfo{booktitle}{\emph{The Semantic Web: 16th International Conference, ESWC 2019, Portoro{\v{z}}, Slovenia, June 2--6, 2019, Proceedings 16}}. Springer, \bibinfo{pages}{459--474}.
\newblock


\bibitem[Ma et~al\mbox{.}(2022)]%
        {ma2022mmekg}
\bibfield{author}{\bibinfo{person}{Yubo Ma}, \bibinfo{person}{Zehao Wang}, \bibinfo{person}{Mukai Li}, \bibinfo{person}{Yixin Cao}, \bibinfo{person}{Meiqi Chen}, \bibinfo{person}{Xinze Li}, \bibinfo{person}{Wenqi Sun}, \bibinfo{person}{Kunquan Deng}, \bibinfo{person}{Kun Wang}, \bibinfo{person}{Aixin Sun}, {and} \bibinfo{person}{Jing Shao}.} \bibinfo{year}{2022}\natexlab{}.
\newblock \showarticletitle{{MMEKG}: Multi-modal Event Knowledge Graph towards Universal Representation across Modalities}. In \bibinfo{booktitle}{\emph{Proceedings of the 60th Annual Meeting of the Association for Computational Linguistics: System Demonstrations}}. \bibinfo{publisher}{Association for Computational Linguistics}, \bibinfo{address}{Dublin, Ireland}, \bibinfo{pages}{231--239}.
\newblock
\urldef\tempurl%
\url{https://doi.org/10.18653/v1/2022.acl-demo.23}
\showDOI{\tempurl}


\bibitem[McInnes et~al\mbox{.}(2018)]%
        {mcinnes2018umap}
\bibfield{author}{\bibinfo{person}{Leland McInnes}, \bibinfo{person}{John Healy}, {and} \bibinfo{person}{James Melville}.} \bibinfo{year}{2018}\natexlab{}.
\newblock \showarticletitle{Umap: Uniform manifold approximation and projection for dimension reduction}.
\newblock \bibinfo{journal}{\emph{arXiv preprint arXiv:1802.03426}} (\bibinfo{year}{2018}).
\newblock


\bibitem[O{\~n}oro-Rubio et~al\mbox{.}(2017)]%
        {onoro2017imagegraph}
\bibfield{author}{\bibinfo{person}{Daniel O{\~n}oro-Rubio}, \bibinfo{person}{Mathias Niepert}, \bibinfo{person}{Alberto Garc{\'\i}a-Dur{\'a}n}, \bibinfo{person}{Roberto Gonz{\'a}lez}, {and} \bibinfo{person}{Roberto~J L{\'o}pez-Sastre}.} \bibinfo{year}{2017}\natexlab{}.
\newblock \showarticletitle{Answering visual-relational queries in web-extracted knowledge graphs}.
\newblock \bibinfo{journal}{\emph{arXiv preprint arXiv:1709.02314}} (\bibinfo{year}{2017}).
\newblock


\bibitem[Suchanek et~al\mbox{.}(2007)]%
        {suchanek2007yago}
\bibfield{author}{\bibinfo{person}{Fabian~M Suchanek}, \bibinfo{person}{Gjergji Kasneci}, {and} \bibinfo{person}{Gerhard Weikum}.} \bibinfo{year}{2007}\natexlab{}.
\newblock \showarticletitle{Yago: a core of semantic knowledge}. In \bibinfo{booktitle}{\emph{Proceedings of the 16th international conference on World Wide Web}}. \bibinfo{pages}{697--706}.
\newblock


\bibitem[Sun et~al\mbox{.}(2020)]%
        {sun2020multi}
\bibfield{author}{\bibinfo{person}{Rui Sun}, \bibinfo{person}{Xuezhi Cao}, \bibinfo{person}{Yan Zhao}, \bibinfo{person}{Junchen Wan}, \bibinfo{person}{Kun Zhou}, \bibinfo{person}{Fuzheng Zhang}, \bibinfo{person}{Zhongyuan Wang}, {and} \bibinfo{person}{Kai Zheng}.} \bibinfo{year}{2020}\natexlab{}.
\newblock \showarticletitle{Multi-modal Knowledge Graphs for Recommender Systems}. In \bibinfo{booktitle}{\emph{Proceedings of the 29th ACM International Conference on Information \& Knowledge Management}} (Virtual Event, Ireland) \emph{(\bibinfo{series}{CIKM '20})}. \bibinfo{publisher}{Association for Computing Machinery}, \bibinfo{address}{New York, NY, USA}, \bibinfo{pages}{1405–1414}.
\newblock
\showISBNx{9781450368599}
\urldef\tempurl%
\url{https://doi.org/10.1145/3340531.3411947}
\showDOI{\tempurl}


\bibitem[Tao et~al\mbox{.}(2021)]%
        {tao2021multi}
\bibfield{author}{\bibinfo{person}{Shaohua Tao}, \bibinfo{person}{Runhe Qiu}, \bibinfo{person}{Yuan Ping}, {and} \bibinfo{person}{Hui Ma}.} \bibinfo{year}{2021}\natexlab{}.
\newblock \showarticletitle{Multi-modal knowledge-aware reinforcement learning network for explainable recommendation}.
\newblock \bibinfo{journal}{\emph{Knowledge-Based Systems}}  \bibinfo{volume}{227} (\bibinfo{year}{2021}), \bibinfo{pages}{107217}.
\newblock


\bibitem[Vrande{\v{c}}i{\'c} and Kr{\"o}tzsch(2014)]%
        {vrandevcic2014wikidata}
\bibfield{author}{\bibinfo{person}{Denny Vrande{\v{c}}i{\'c}} {and} \bibinfo{person}{Markus Kr{\"o}tzsch}.} \bibinfo{year}{2014}\natexlab{}.
\newblock \showarticletitle{Wikidata: a free collaborative knowledgebase}.
\newblock \bibinfo{journal}{\emph{Commun. ACM}} \bibinfo{volume}{57}, \bibinfo{number}{10} (\bibinfo{year}{2014}), \bibinfo{pages}{78--85}.
\newblock


\bibitem[Wang et~al\mbox{.}(2020)]%
        {wang2020richpedia}
\bibfield{author}{\bibinfo{person}{Meng Wang}, \bibinfo{person}{Haofen Wang}, \bibinfo{person}{Guilin Qi}, {and} \bibinfo{person}{Qiushuo Zheng}.} \bibinfo{year}{2020}\natexlab{}.
\newblock \showarticletitle{Richpedia: a large-scale, comprehensive multi-modal knowledge graph}.
\newblock \bibinfo{journal}{\emph{Big Data Research}}  \bibinfo{volume}{22} (\bibinfo{year}{2020}), \bibinfo{pages}{100159}.
\newblock


\bibitem[Wen et~al\mbox{.}(2021)]%
        {wen2021resin}
\bibfield{author}{\bibinfo{person}{Haoyang Wen}, \bibinfo{person}{Ying Lin}, \bibinfo{person}{Tuan Lai}, \bibinfo{person}{Xiaoman Pan}, \bibinfo{person}{Sha Li}, \bibinfo{person}{Xudong Lin}, \bibinfo{person}{Ben Zhou}, \bibinfo{person}{Manling Li}, \bibinfo{person}{Haoyu Wang}, \bibinfo{person}{Hongming Zhang}, {et~al\mbox{.}}} \bibinfo{year}{2021}\natexlab{}.
\newblock \showarticletitle{Resin: A dockerized schema-guided cross-document cross-lingual cross-media information extraction and event tracking system}. In \bibinfo{booktitle}{\emph{Proceedings of the 2021 Conference of the North American Chapter of the Association for Computational Linguistics: Human Language Technologies: Demonstrations}}. \bibinfo{pages}{133--143}.
\newblock


\bibitem[Yu et~al\mbox{.}(2020)]%
        {yu2020cross}
\bibfield{author}{\bibinfo{person}{Jing Yu}, \bibinfo{person}{Zihao Zhu}, \bibinfo{person}{Yujing Wang}, \bibinfo{person}{Weifeng Zhang}, \bibinfo{person}{Yue Hu}, {and} \bibinfo{person}{Jianlong Tan}.} \bibinfo{year}{2020}\natexlab{}.
\newblock \showarticletitle{Cross-modal knowledge reasoning for knowledge-based visual question answering}.
\newblock \bibinfo{journal}{\emph{Pattern Recognition}}  \bibinfo{volume}{108} (\bibinfo{year}{2020}), \bibinfo{pages}{107563}.
\newblock


\end{thebibliography}

\end{document}